\documentclass{article}
\usepackage{kr}

\usepackage{times}
\usepackage{soul}
\usepackage{url}
\usepackage[hidelinks]{hyperref}
\usepackage[utf8]{inputenc}
\usepackage[small]{caption}
\usepackage{graphicx}
\usepackage{amsmath}
\usepackage{amsthm}
\usepackage{booktabs}
\usepackage{algorithm}
\usepackage{algorithmic}
\usepackage{CJKutf8}

\usepackage{array}
\usepackage{listings}
\usepackage{xcolor}
\usepackage{multirow}

\colorlet{punct}{red!60!black}
\definecolor{background}{HTML}{EEEEEE}
\definecolor{delim}{RGB}{20,105,176}
\colorlet{numb}{magenta!60!black}
\lstdefinelanguage{json}{
    basicstyle=\normalfont\ttfamily,
    numbers=left,
    numberstyle=\scriptsize,
    stepnumber=1,
    numbersep=3pt,
    showstringspaces=false,
    breaklines=true,
    frame=lines,
    backgroundcolor=\color{background},
    literate=
     *{0}{{{\color{numb}0}}}{1}
      {1}{{{\color{numb}1}}}{1}
      {2}{{{\color{numb}2}}}{1}
      {3}{{{\color{numb}3}}}{1}
      {4}{{{\color{numb}4}}}{1}
      {5}{{{\color{numb}5}}}{1}
      {6}{{{\color{numb}6}}}{1}
      {7}{{{\color{numb}7}}}{1}
      {8}{{{\color{numb}8}}}{1}
      {9}{{{\color{numb}9}}}{1}
      {:}{{{\color{punct}{:}}}}{1}
      {,}{{{\color{punct}{,}}}}{1}
      {\{}{{{\color{delim}{\{}}}}{1}
      {\}}{{{\color{delim}{\}}}}}{1}
      {[}{{{\color{delim}{[}}}}{1}
      {]}{{{\color{delim}{]}}}}{1},
}

\title{Exploiting LLMs' Reasoning Capability to Infer Implicit Concepts in Legal Information Retrieval}

\author{%
Hai-Long Nguyen$^1$\and
Tan-Minh Nguyen$^1$\and
Duc-Minh Nguyen$^2$\\
Thi-Hai-Yen Vuong$^1$ \and
Ha-Thanh Nguyen$^3$ \and
Xuan-Hieu Phan$^1$ \\
Ken Satoh$^3$
\affiliations
$^1$ VNU University of Engineering and Technology, Hanoi\\
$^2$ RMIT University Vietnam\\
$^3$ National Institute of Informatics, Japan\\
\emails
long.nh@vnu.edu.vn
}

\begin{document}

\maketitle

\begin{abstract}
  
Statutory law retrieval is a typical problem in legal language processing, that has various practical applications in law engineering. Modern deep learning-based retrieval methods have achieved significant results for this problem. However, retrieval systems relying on semantic and lexical correlations often exhibit limitations, particularly when handling queries that involve real-life scenarios, or use the vocabulary that is not specific to the legal domain. In this work, we focus on overcoming this weaknesses by utilizing the logical reasoning capabilities of large language models (LLMs) to identify relevant legal terms and facts related to the situation mentioned in the query. The proposed retrieval system integrates additional information from the term--based expansion and query reformulation to improve the retrieval accuracy. The experiments on COLIEE 2022 and COLIEE 2023 datasets show that extra knowledge from LLMs helps to improve the retrieval result of both lexical and semantic ranking models. The final ensemble retrieval system outperformed the highest results among all participating teams in the COLIEE 2022 and 2023 competitions.

\end{abstract}

\section{Introduction} 
\label{sec:introduction}

The statute law retrieval problem is a typical task in Legal NLP. The problem takes as input a natural language query, which could be a question, legal statement, or a specific scenario. The output of the problem is relevant legal articles or segments extracted from articles containing information that address the given query. In countries that follow the statute law system, this problem is the vital core of legal reference systems or legal search engines which could serve for many people, from legal experts to non-experts. These AI-based tools could help legal practitioners reduce the amount of time and resources on paperwork. 

Previous studies have done well in using lexical features through statistical models combined with semantic features from Transformer-based models to address and improve retrieval results \cite{kim2022coliee,goebel2023summary,louis2022statutory}.
Given the high number of candidates, the legal retrieval task is often split into two steps: lexical ranking and semantic re--ranking. However, a significant issue arises when queries containing content with no lexical overlap with the gold standard articles, which leads to the gold standard articles being eliminated from the first retrieval phase. 
Moreover, different query types exist, including legal statements and specific scenarios as shown in Table \ref{tab:data-example}. Legal statements are typically concise and may not require logical reasoning for comprehension. Specific scenarios describe conflicts of rights between parties, involving more complex underlying logical reasoning.

\begin{CJK}{UTF8}{min}
\begin{table*}[ht]
\caption{An example of a specific scenario query in the COLIEE dataset.}
\label{tab:data-example}
\begin{tabular}{lp{7cm}p{7cm}}
\hline
& \multicolumn{1}{c}{\textbf{Japanese}}  & \multicolumn{1}{c}{\textbf{English}}  \\ \hline
\textbf{Query}             & Ａが自己所有の動産甲をＢに賃貸し引き渡していた場合において、ＣがＢのもとから甲を窃取したときは、Ａは、Ｃに対して、占有回収の訴えによって甲の返還を求めることができる。  & A has leased and delivered movable X owned by A to B. If C steals X from B, A may demand C to return X by an action for recovery of possession.            \\ \hline
\textbf{LLMs generation}   &    &  \\ 
Legal terms                & 占有回収の訴え     & Action for recovery of possession                                         \\ 
Re-write query             & このクエリに関連する法的概念は「占有回収訴訟」です。占有回収訴訟とは、動産を不法に占有する者に対して、その返還を求める訴訟です。本件では、CがBのもとから甲を窃取した場合、AはCに対して占有回収訴訟によって甲の返還を求めることができます。なぜなら、Aは甲の所有者であり、Cは甲を不法に占有しているからです。ただし、例外として、善意かつ無過失で動産を取得した者は、たとえその動産が盗品であったとしても、占有回収訴訟によって返還を求められることはありません（善意取得）。 & The legal concept relevant to this query is “possession recovery litigation.” A possession recovery lawsuit is a lawsuit that seeks the return of movable property against a person who illegally possesses it. In this case, if C steals Party A from B, A can demand the return of Party A by filing a possession recovery lawsuit against C. This is because A is the owner of Party A, and C is in illegal possession of Party A. However, as an exception, a person who acquires movable property in good faith and without fault will not be required to return it through a possession recovery lawsuit, even if the movable property is stolen (good faith acquisition). \\ \hline
\textbf{Relevant articles} & 第百八十一条 占有権は、代理人によって取得することができる。 & Possessory rights may be acquired through an agent. \\ \hline
\end{tabular}
\end{table*}
\end{CJK}

The emergence of Large Language Models (LLMs) has opened up a new era of development in the field of AI, especially NLP \cite{zhao2023survey,yang2023harnessing,bommasani2021opportunities}. To date, LLMs have effectively addressed various general NLP tasks, including text summarization and machine translation
\cite{huang2023towards,yao2023empowering,laban2023summedits}. For legal domains, these models have shown their abilities to improve the performance of downstream tasks \cite{yu2022legal,trautmann2022legal,zhou2023boosting}. However, previous work relied on prompting techniques or the in-context learning capability of LLMs to tackle these tasks straightforwardly. In contrast, this work leverages the strength of LLMs to explore hidden logical reasoning of queries as additional information to retrieval models. 



The proposed method is inspired by how legal experts search legal documents. When searching for legal documents that relate to a real-life situation, legal experts typically begin by finding which behaviours the subjects perform, which legal issues are described in the situation, and which legal concepts they relate to. Then, they search for legal documents related to those legal issues or concepts. Identifying relevant legal issues and concepts related to real-life situations is equivalent to the process of query expansion in terms of computer science. However, this process is not merely semantic matching but requires inference based on the interaction between entities in the event. Recent studies have shown that large language models such as GPT-4 \cite{achiam2023gpt} and Gemini have basic inference capabilities, although they have not reached human-level inference \cite{achiam2023gpt,rane2024gemini,wang2024can}. This has motivated this research to utilize LLMs to explore hidden logical semantics of queries for query expansion.

The main contributions of this research include identifying the current weaknesses of lexical and semantic feature-based matching models and proposing two query expansion methods using LLMs to extract the underlying topic of that query. The term generation  and query-reformulation techniques are used and injected in both lexical and semantic ranking models. The proposed query expansion methods enable the retrieval system to effectively surpass the top-performing approaches of the COLIEE 2022 and 2023 datasets. 

The paper will be structured as follows: the next section presents some related studies on the statute law retrieval problem and the application of LLMs to this problem, section \ref{sec:query-expan-method} presents proposed method and experiments. Finally, the section \ref{sec:conclusion} concludes the paper and highlights some future work.

\section{Related Work}
Legal NLP has attracted much interest from researchers and companies because of its potential and wide range of applications \cite{chalkidis2019deep,zhong2020does}. Statutory article retrieval is one of the core problems in this field, playing an essential role in search engines and reference systems. A traditional approach for query-article matching is the term--based model \cite{10.1145/3322640.3326742,10.1145/3322640.3326740}. However, the trend rapidly changes to deep learning models involving word embedding \cite{landthaler2016extending,kayalvizhi2019legal}, document vector embedding \cite{sugathadasa2019legal}, and contextual understanding \cite{nguyen2024captain,nguyen2023nowj1}. Recently, a growing number of studies have focused on large language models (LLMs) in legal NLP \cite{sun2023short,lai2023large}. These models can address downstream tasks such as question answering \cite{yu2022legal}, legal judgment prediction \cite{trautmann2022legal}, legal summarization \cite{pont2023legal,gesnouin2024llamandement} based on fine-tuning and prompt engineering techniques. However, there is space for exploiting LLMs to address legal information retrieval. Zhou et al. \cite{zhou2023boosting} utilized LLMs to extract significant content from legal cases and incorporate it into retrieval models. Unlike \cite{zhou2023boosting}, which involved a manual annotation process with the assistance of legal experts to collect salient content, this work does not rely on human-labeled data for LLMs. This distinction enhances the reproducibility and openness of the research.

Query expansion is one of the typical methods to shorten the gap between query and document, further improving retrieval performance. The expansion process mainly relies on external knowledge-based and pseudo-relevance feedback to enrich query information. Relevance documents can serve as additional knowledge in query expansion. However, this approach does not apply to our work as the relevant legal documents are not provided with the query. Therefore, we generate legal terms and a new query based on the original query to serve as additional knowledge for the retrieval system. Recently studies have leveraged generative models to rewrite the query via fine-tuning with labeled data \cite{imani2019deep,zheng2020bert,10.1016/j.ipm.2021.102672}. Large language models (LLMs) \cite{team2023gemini,touvron2023llama,achiam2023gpt} with billions of parameters are trained on massive volumes of data. 
LLMs could address downstream tasks via prompt engineering or instructions. Wang et al. \cite{wang2023query2doc} utilized \textit{gpt-3.5-turbo} to generate a pseudo document from a given query using few-shot prompting. Jagerman et al. \cite{jagerman2023query} studied the performance of various prompting techniques in query expansion using Flan-family models. These works used common information retrieval datasets such as MS-MACRO \cite{DBLP:conf/nips/NguyenRSGTMD16} and BEIR \cite{thakur2021beir} as evaluation data, proving the effectiveness of LLMs-based query expansion. This paper focuses on inferring implicit concepts based on a given legal query. To the best of our knowledge, we are the first to exploit general LLMs as a query rewriter in the legal statute law retrieval problem.


\section{Query expansion for legal document retrieval}
\label{sec:query-expan-method}

As mentioned in section \ref{sec:introduction}, to address complex queries with multiple layers of implicit semantics, reasoning capabilities are required. This section will describe the prompting patterns on Large Language Models (LLMs) to leverage their basic reasoning abilities and extract key information. Additionally, the legal document retrieval system described in section \ref{subsec:retrieval-system} will integrate ranking results from multiple models and is designed to entirely leverage information obtained from the prompting process.

\subsection{LLMs-based legal term extraction}
\label{subsec:legal-term-extraction}
Due to theirs generation capability and knowledge across various domains, LLMs are utilized to generate the key terms which represents the general topics of the query. 
Legal terms related to the query are extracted using zero--shot prompting techniques on LLMs. Since the experimented dataset is the COLIEE dataset with the original version written in Japanese, the prompting sentence will be written in Japanese to avoid information loss. Furthermore, to facilitate the processing of LLMs' output, the prompting sentence includes additional instructions for LLMs output in JSON format. The prompting pattern and instruction in English version are described in listing \ref{lst:term-prompt-en}, the one with Japanese version is presented at the listing  \ref{lst:term-prompt-jp} in the Appendix.


\begin{lstlisting}[label={lst:term-prompt-en}, language=json,firstnumber=1,  basicstyle=\small, caption=Prompting pattern for legal term extraction\\ (English version), escapechar=\#]
#Given a legal situation, find the relevant facts and legal concepts#
#relevant to that situation#:{query}

#The output must be formatted as a JSON instance conforming to# 
#the JSON schema below. For example, the schema#
{
  properties: {
    foo: {
      title: Foo, 
      description: a list of strings, 
      type: array, 
      items: {type: string}
    }}, 
  required: [foo]}
\end{lstlisting}

To simplify the experiments and verify the impact of the expanded information, the collection of legal terms will be directly concatenated with the query at the lexical layer. Since this concatenation disrupts the semantic integrity at the end of the query, lexical-based retrieval models will be most appropriate for this query--expansion type. The BM25 \cite{robertson2009probabilistic} model which is one of the most commonly lexical-based ranking models is utilized. The process of using BM25 to rank this type of expanded queries will be described in section \ref{subsec:retrieval-system}.

\subsection{LLMs-based query augmentation}
\label{subsec:query-aug}

Directly concatenating legal terms into the query does not ensure semantic coherence for the query, which may cause confusion for semantic--based ranking models. To preserve the semantic integrity, the legal--style oriented query reformulation is performed by zero--shot prompting technique on LMMs. The suggested prompting pattern written in English is detailed in listing \ref{lst:redesc-prompt-en}, the Japanese one is showed at the listing \ref{lst:redesc-prompt-jp}.

\begin{CJK}{UTF8}{min}
\begin{lstlisting}[label={lst:redesc-prompt-en}, language=json,firstnumber=1,  basicstyle=\small, caption=Prompting pattern for redescribed query \\ (English version), escapechar=\#]
#Given a legal situation, extract the relevant facts and legal# 
#concepts relevant to that situation#:{query}
\end{lstlisting}
\end{CJK}

The set of reformulated queries will be used to train the semantic matching model, further supporting information about the legal speciality of the content mentioned in the original query. The specific method to leverage the prompting results from LLMs will be described in section \ref{subsec:retrieval-system}. 

\subsection{Retrieval system utilizing expanded query}
\label{subsec:retrieval-system}

\begin{figure}[ht]
\centering
\includegraphics[width=0.4\textwidth]{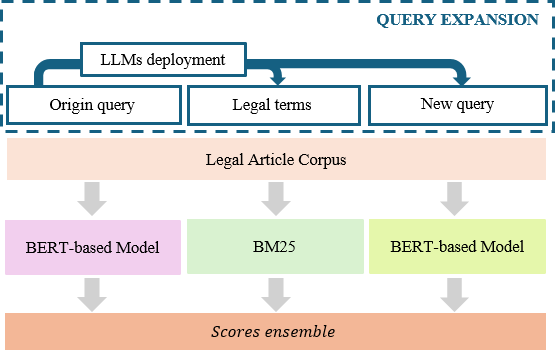}
\caption{Query expansion supported retrieval system}
\label{fig:retrieval-system-1}
\end{figure}


To combine the information from both prompting methods described in section \ref{subsec:legal-term-extraction} and \ref{subsec:query-aug}, both lexical-based ranking and semantic-based ranking models will be utilized and ensembled.

Assuming the input query is denoted as $\mathbf{q}$ and the legal corpus, denoted as set $\mathbf{D}$, including $m$ legal documents: $\mathbf{D} = \{d_1, d_2, ..., d_m\}$. After the legal-term extraction, the obtained terms form the set $T_q = \{t_1, t_2, ..., t_n$\}. These terms are concatenated with the query at the lexical level to create a new query:
\begin{equation*}
    \mathbf{q}_\text{term-expand} = \text{concat}(q, t_1, t_2, ..., t_n)
\end{equation*}
The BM25 model is employed to calculate the lexical relevance between the expanded query $\mathbf{q}_\text{term-expand}$ and the legal document $d_i$. The relevance score calculated by the BM25 model is denoted as: 
\begin{equation*}
    \mathbf{R}_\text{BM25} = \text{BM25}(q_\text{term-expand}, d_i)
\end{equation*}
For the semantic-based ranking component, we utilized ranking models based on the BERT architecture, fine-tuned by sequence classification tasks. Experimenting with the simplest backbone model such as BERT will help assess the contribution of the additional information from the query expansion process. Two distinct BERT models are employed in this proposed retrieval system, one for ranking the original query and the other for the reformulated query. These two BERT models are described as follows:
\begin{itemize}
    \item Let $\mathbf{q}_\text{origin}$ be the original query. The first semantic ranking model denoted as $\text{BERT}_\text{origin}$, evaluates the semantic relevance between the original query $\mathbf{q}_{\text{origin}}$ and the legal document $d_i$, denoted as: \begin{equation*}
        \mathbf{R}_{ori} = \text{BERT}_{\text{origin}}(\mathbf{q}_{\text{origin}}, d_i)
    \end{equation*}
    \item Let $\mathbf{q}_{\text{reformulated}}$ be the reformulated query, derived from LLMs. The second BERT model, denoted as $\text{BERT}_{\text{reformulated}}$, evaluates the semantic relevance between the reformulated query $\mathbf{q}_{\text{reformulated}}$ and the legal document $d_i$, denoted as: \begin{equation*}
        \mathbf{R}_{\text{reform}} = \text{BERT}_{\text{reformulate}}(\mathbf{q}_{\text{reformulated}}, d_i)
    \end{equation*}
\end{itemize}
The architecture of the retrieval system, which includes three ranking models, is illustrated in Figure \ref{fig:retrieval-system-1}. The final relevance scores from all three ranking models will be weighted ensembled using the equation \ref{eq:weighted-ensemble}.
\begin{align}
    \label{eq:weighted-ensemble}
    \centering
    \begin{split}
        \vspace{-0.3cm}
        R_{\textit{final}} &= \alpha * \mathbf{R}_{\text{ori}} + \beta * \mathbf{R}_{\text{bm25}} + \gamma * \mathbf{R}_{\text{reform}} \\
        s.t: \alpha + \beta + \gamma &= 1
    \end{split}
\end{align}
where:
\begin{itemize}
    \item $\mathbf{R}_{\text{ori}}$ represents the correlation score between the original query and the article as calculated by the BERT-based ranking model.
    \item $\mathbf{R}_{\text{bm25}}$ is the correlation score between the query, which has been concatenated with legal term outputs from LLMs, and the article, as calculated by the BM25 ranking algorithm.
    \item $\mathbf{R}_{\text{reform}}$ is the correlation score between the query that has been re-described using LLMs and the article, also calculated by the BERT-based ranking model.
    \item $\alpha$, $\beta$, and $\gamma$ are the weights assigned to the BERT model with the original query, the BM25 model with the legal--term expanded query, and the BERT model with the re-described query, respectively. The optimal values of these three weights are selected through a grid-search process on a validation set.
\end{itemize}
    
\subsection{Inference strategy}
\label{subsec:inference-strategy}

\begin{figure}[ht]
\centering
\includegraphics[width=0.4\textwidth]{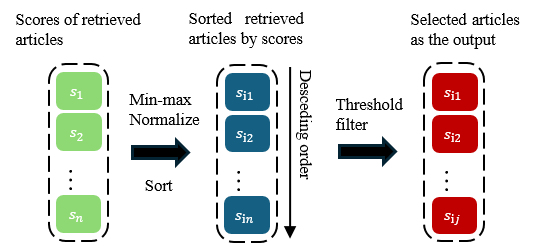}
\caption{Retrieval task post-processing}
\label{fig:post-processing}
\end{figure}

The legal document retrieval task is indeed about ranking the documents in the law corpus based on the relevancy between each document and the input query. 
Meanwhile, the training process for the BERT model uses the objective function of the Sequence Classification downstream task, classifying the \textit{(query-article)} pair into \textit{positive} or \textit{negative} labels. Therefore, directly using the relevance score from the BERT model often does not yield the highest effectiveness for the retrieval system. To address this drawback, after obtaining the $R_{\textit{final}}$ for each \textit{(query-article)} pair, a post-processing phase will be conducted. In the post-processing phase, the relevance scores for all \textit{(query-article)} pairs corresponding to the same query will be min-max normalized. Subsequently, a common optimal threshold will be determined through a grid-search process on the validation set. All articles with a relevance score exceeding this threshold will be selected as relevant to the query. Specifically, the post-processing procedure for producing the final set of relevant documents is described in Figure \ref{fig:post-processing}.

\section{Experiment and result}

\subsection{COLIEE dataset introduction}

To evaluate the effectiveness of the proposed retrieval system, the COLIEE 2022 \footnote{\url{https://sites.ualberta.ca/~rabelo/COLIEE2022/}} and COLIEE 2023 \footnote{\url{https://sites.ualberta.ca/~rabelo/COLIEE2023/}} datasets for the Statute Law Retrieval task will be utilized. Both datasets employ a corpus derived from the Japanese statute law, containing $782$ legal provisions. 
\begin{table}[ht]
\caption{Statute law corpus token statistics}
\centering
\begin{tabular}{lll}
\hline
\multicolumn{1}{c}{\textbf{}} & \multicolumn{2}{l}{\textbf{\#word per articles}} \\ \hline
                              & \textbf{en}             & \textbf{jp}            \\ \hline
\textbf{Min}                  & 1                       & 6                      \\ \hline
\textbf{Max}                  & 655                     & 755                    \\ \hline
\textbf{Average}              & 71.83                   & 93.34                  \\ \hline
\end{tabular}
\label{tab:corpus-statistic}
\end{table}
\begin{table}[ht]
\caption{Token statistics}
\begin{tabular}{lccccc}
\hline
                                  & \multicolumn{1}{l}{\textbf{}} & \multicolumn{2}{l}{\textbf{COLIEE 2022}} & \multicolumn{2}{l}{\textbf{COLIEE 2023}} \\ \hline
\textbf{\#sample}                 & -                                      & 887                 & 109                 & 996                 & 101                 \\ \hline
\multirow{2}{*}{\textbf{Min}}     & EN                                     & 6                   & 13                  & 6                   & 11                  \\
                                  & JP                                     & 10                  & 18                  & 10                  & 15                  \\ \hline
\multirow{2}{*}{\textbf{Max}}     & EN                                     & 149                 & 91                  & 149                 & 27                  \\
                                  & JP                                     & 202                 & 125                 & 202                 & 109                 \\ \hline
\multirow{2}{*}{\textbf{Average}} & EN                                     & 39                  & 42                  & 39                  & 42                  \\
                                  & JP                                     & 51                  & 52                  & 51                  & 54                  \\ 
\end{tabular}
\end{table}
The labeled set of COLIEE 2022 and COLIEE 2023 respectively includes $887$ and $996$ instances. The content of queries in both datasets often describes a legal statement, legal situation or a real--life legal scenario. Table 2 summarizes the number of samples and the length of each query through the statistical indices \textit{minimum}, \textit{maximum}, and \textit{average}.

\subsection{Implementation detail}

In this section, the implementation, training, inference, and ensemble of model results are discussed. 
Firstly, for the BM25 model, we utilize the rank-bm25 library\footnote{\url{https://github.com/dorianbrown/rank_bm25}}. Before being processed by the BM25 model, articles and queries written in Japanese are word--based tokenized using the Konoha library \footnote{\url{https://github.com/himkt/konoha?tab=readme-ov-file}} and the MeCab tokenizer. The prompting process is performed on GPT--4 \cite{achiam2023gpt}  and Gemini \cite{team2023gemini}. 

\begin{table}
\caption{Recall score of BM25 model in Japanese--version training dataset of COLIEE 2022 and COLIEE 2023 with original query and term--expanded query using two LLMs models: Gemini and GPT--4.}
\begin{tabular}{cclcc}
\hline
\multicolumn{1}{l}{\textbf{Datasets}}                                            & \multicolumn{1}{l}{\textbf{Top-k}} & \textbf{\begin{tabular}[c]{@{}l@{}}Origin\end{tabular}} & \multicolumn{1}{l}{\textbf{\begin{tabular}[c]{@{}l@{}}Gemini\end{tabular}}} & \multicolumn{1}{l}{\textbf{\begin{tabular}[c]{@{}l@{}}GPT--4\end{tabular}}} \\ \hline
\multirow{5}{*}{\textbf{\begin{tabular}[c]{@{}c@{}}COLIEE \\ 2022\end{tabular}}} & 30                                 & 0.7590                                                             & 0.8423                                                                                  & 0.8394                                                                                 \\
                                                                                 & 50                                 & 0.7878                                                             & 0.8735                                                                                  & 0.8663                                                                                 \\
                                                                                 & 100                                & 0.8394                                                             & 0.9098                                                                                  & 0.9061                                                                                 \\
                                                                                 & 200                                & 0.8836                                                             & 0.9432                                                                                  & 0.9366                                                                                 \\
                                                                                 & 500                                & \textbf{0.9418}                                                             & \textbf{0.9782}                                                                                  & \textbf{0.9776}                                                                                 \\ \hline
\multirow{5}{*}{\textbf{\begin{tabular}[c]{@{}c@{}}COLIEE \\ 2023\end{tabular}}} & 30                                 & 0.8320                                                             & 0.8530                                                                                  & 0.8516                                                                                 \\
                                                                                 & 50                                 & 0.8592                                                             & 0.8834                                                                                  & 0.8767                                                                                 \\
                                                                                 & 100                                & 0.8955                                                             & 0.9182                                                                                  & 0.9139                                                                                 \\
                                                                                 & 200                                & 0.9316                                                             & 0.9479                                                                                  & 0.9421                                                                                 \\
                                                                                 & 500                                & \textbf{0.9736}                                                             & \textbf{0.9790}                                                                                  & \textbf{0.9785}                                                                                 \\ \hline
\end{tabular}
\label{tab:recall-score-bm25}
\end{table}
The training dataset provides sample which contains a query and a list of gold relevant articles. For training the BERT-based ranking model, the training samples are transformed into pairwise training samples. The transformed set contains a query and an article with two labels $0$ or $1$ representing not--relevant or relevant. The negative samples are chosen from top candidates according to the BM25's relevance score. To achieve the optimal ratio between negative and positive samples, the top--30 most relevant candidates according to the BM25 model are utilized for creating negative samples for the queries. The Huggingface  library \footnote{\url{https://huggingface.co}} is employed for creating and training the BERT-based model with the downstream task sequence classification. The pre--trained Multilingual BERT \footnote{\url{https://huggingface.co/google-bert/bert-base-multilingual-cased}} is being used for initializing the BERT-based retrieval model. The experimented models are fine--tuned with the training data in $3$ epochs. There are two ranking BERT-based models in the proposed retrieval system which are the ranking model fine--tuned with the origin query and the ranking model fine--tuned with the re--described query. The results in section \ref{sec:exp-res} will demonstrate that using two separate BERT-based ranking models for original and re-described queries is more effective than a single fine-tuned BERT model handling both.

\begin{table}[ht]
\caption{Experiment results on private test of COLIEE 2022 dataset.}
\centering
\begin{tabular}{lccc}
\hline
\multirow{2}{*}{\textbf{Teams/ Models}}                                       & \multicolumn{3}{c}{\textbf{Private Test}}           \\
                                                                              & \textbf{F2}     & \textbf{P}      & \textbf{R}      \\ \hline
\textbf{HUKB}                                                                 & 0.8200          & \textbf{0.8180} & 0.8405          \\
\textbf{OVGU}                                                                 & 0.7790          & 0.7781          & 0.8054          \\
\textbf{JNLP}                                                                 & 0.7699          & 0.6865          & 0.8378          \\
\textbf{UA}                                                                   & 0.7638          & 0.8073          & 0.7641          \\ \hline
\textbf{\begin{tabular}[c]{@{}l@{}}Proposed \\ Retrieval System\end{tabular}} & \textbf{0.8449} & 0.7935          & \textbf{0.9005} \\ \hline
\end{tabular}
\label{tab:experiment-result-table-2022}
\end{table}

\begin{table}[ht]
\caption{Experiment results on private test of COLIEE 2023 dataset.}
\centering
\begin{tabular}{lccc}
\hline
\multirow{2}{*}{\textbf{Teams/ Models}}                                       & \multicolumn{3}{c}{\textbf{Private Test}}          \\
                                                                              & \textbf{F2}     & \textbf{P}      & \textbf{R}     \\ \hline
\textbf{CAPTAIN}                                                              & 0.757           & 0.726           & 0.792          \\
\textbf{JNLP}                                                                 & 0.745           & 0.645           & \textbf{0.867} \\
\textbf{NOWJ}                                                                 & 0.727           & 0.682           & 0.772          \\
\textbf{HUKB}                                                                 & 0.6473          & 0.6248          & 0.6708         \\ \hline
\textbf{\begin{tabular}[c]{@{}l@{}}Proposed \\ Retrieval System\end{tabular}} & \textbf{0.7601} & \textbf{0.7254} & 0.8366         \\ \hline
\end{tabular}
\label{tab:experiment-result-table-2023}
\end{table}

Before the training phase, a validation set is separated from $20\%$ of the labelled set, and the remaining $80\%$ of samples are formed into the training set. After the training the retrieval model, 
a grid--search process is conducted on the validation set for choosing the best $\alpha$, $\beta$, $\gamma$ and $\textit{threshold}$.

\subsection{Experiment Result}
\label{sec:exp-res}

The recall score of two term--expansion methods are presented in the table \ref{tab:recall-score-bm25}. 
According to that table, the recall scores when using the term--expanded queries instead of the original one are significantly higher at all top--k levels, with both Gemini and GPT--4 terms--expansion. This proves that adding legal terms will make the queries more lexically relevant to the gold standard articles, facilitating the BM25 model to rank more accurately. 
Additionally, the recall score of the Gemini--based term--expansion achieved a higher recall score than the GPT--4. Therefore, the BM25's relevance score with Gemini's expanded queries is used in the final retrieval system.

\begin{table*}[ht]
\caption{\small Evaluation of some retrieval system's variations. The $\text{BM25}_\text{legalterm}$ is the BM25 model worked with the term--expanded query instead of original query.}
\centering
\begin{tabular}{lcccccc}
\hline
\multirow{2}{*}{\textbf{\begin{tabular}[c]{@{}l@{}}Retrieval system's variations\end{tabular}}} & \multicolumn{3}{c}{\textbf{2022 Private Test}}               & \multicolumn{3}{c}{\textbf{2023 Private Test}}      \\
                                                                                                   & \textbf{F2}     & \textbf{P}      & \textbf{R}      & \textbf{F2}     & \textbf{P}      & \textbf{R}      \\ \hline
\begin{tabular}[c]{@{}l@{}}$\text{BM25}_\text{legalterm}$ + $\text{BERT}_\text{origin} $\end{tabular}                     & 0.8305          & \textbf{0.8107} & 0.8680          & 0.7356          & 0.7227          & 0.7871          \\
\hline
\begin{tabular}[c]{@{}l@{}}$\text{BM25}_\text{legalterm}$ + $\text{BERT}_\text{reform}$\end{tabular}                 & 0.8201          & 0.7928          & 0.8515          & 0.7050          & 0.6787          & 0.7574          \\ \hline
\begin{tabular}[c]{@{}l@{}}$\text{BM25}_\text{legalterm}$ + $\text{BERT}_{\text{origin}}$ + $\text{BERT}_\text{reform}$\end{tabular} & \textbf{0.8449} & 0.7953          & \textbf{0.9005} & \textbf{0.7643} & \textbf{0.7478} & \textbf{0.8218} \\ \hline
\end{tabular}
\label{tab:experiment-result-variation-2023}
\end{table*}

According to the implementation detail, the retrieval system is experimented with the COLIEE 2022 and 2023 datasets. Because the BERT's training sample pair is generated by filtering from BM25's candidates, the relevance score derived from BERT--based ranking model need to be ensembled with the relevance score from BM25 model to achieve full potential. The table \ref{tab:experiment-result-variation-2023} shows the results of three variants of retrieval system including: 
\begin{itemize}
    \item (1) The system that ensemble BM25 model with term-expanded queries and the BERT model with original queries.
    \item (2) The system that ensemble BM25 model with term-expanded queries and the BERT model with reformulated queries.
    \item (3) The system that ensemble BM25 model with term-expanded queries, BERT model with original queries and another BERT model with reformulated queries.
\end{itemize}
The retrieval performance of the system (2) is lower than system (1). This decrease occurs because using LLMs to reformulated queries make the query's distribution be changed significantly. However, the combination of three models achieved the highest F2--score which is $0.8449$ for the COLIEE 2022 dataset and $0.7643$ for the COLIEE 2023 dataset. This observation suggests that despite having different distribution with the original query, the reformulated query contributes different--perspective information for the retrieval system. 

The table \ref{tab:experiment-result-table-2022} (COLIEE 2022) and table \ref{tab:experiment-result-table-2023} (COLIEE 2023) show the F2, precision and recall of the retrieval system (3) and results of all teams that participated the competition that years.
As indicated on the table, the proposed retrieval system achieved an F2 score that was $2.49\%$ higher than the best-performing team in COLIEE 2022, and $0.3\%$ higher than CAPTAIN team (best performance team) in COLIEE 2023. This result demonstrates the effectiveness of retrieval system which utilizes legal-specific knowledge extracted from LLMs and the way for integrating the knowledge into the retrieval system.

\section{Conclusion}
\label{sec:conclusion}
In this study, two query expansion methods including legal-term extraction and query reformulation are used to exploiting the reasoning capability and general knowledge from LLMs. The information extracted from these two prompting method are injected into lexical--based and semantic--based ranking models. The experiments in COLIEE 2022 and 2023 dataset show that the generated legal information from LLMs  significantly improves retrieval results and outperforms the results of the teams participating in the COLIEE competition in 2022 and 2023.

\section*{Acknowledgments}
Hai-Long Nguyen was funded by the Master, PhD Scholarship Programme of
Vingroup Innovation Foundation (VINIF), code VINIF.2023.ThS.075.

\section*{Appendix}
\appendix

Due to the original data is written in Japanese, the prompting pattern used will be in Japanese. Listing \ref{lst:term-prompt-jp} describes the prompting pattern of the legal term extraction process and listing \ref{lst:redesc-prompt-jp} shows the prompting pattern for generating the re--described query.

\begin{CJK}{UTF8}{ipxm}
\begin{lstlisting}[label={lst:term-prompt-jp}, language=json,firstnumber=1,  basicstyle=\small, caption=Prompting pattern for legal term extraction\\ (Japanese version), escapechar=\#]
#法的な状況が与えられた場合、その状況に関連す#
#る適切な事実と法的概念を抽出します#:{query}

#出力は、以下のJSONスキーマに準拠するJSONイン#
#スタンスとしてフォーマットする必要がありま#
#す。例として、スキーマ #
{
  properties: {
    foo: {
      title: Foo, 
      description: a list of strings, 
      type: array, 
      items: {type: string}
    }}, 
  required: [foo]
}
#の場合、オブジェクト# {foo: [bar, baz]}
\end{lstlisting}
\end{CJK}

\begin{CJK}{UTF8}{min}
\begin{lstlisting}[label={lst:redesc-prompt-jp}, language=json,firstnumber=1,  basicstyle=\small, caption=Prompting pattern for redescribed query \\ (Japanese version), escapechar=\#]
#法的な状況が与えられた場合、その状況に関連す#
#る適切な事実と法的概念を抽出します#:{query}
\end{lstlisting}
\end{CJK}

\bibliographystyle{kr}
\bibliography{kr-sample}

\begin{thebibliography}{}

\bibitem[\protect\citeauthoryear{Achiam \bgroup et al\mbox.\egroup }{2023}]{achiam2023gpt}
Achiam, J.; Adler, S.; Agarwal, S.; Ahmad, L.; Akkaya, I.; Aleman, F.~L.; Almeida, D.; Altenschmidt, J.; Altman, S.; Anadkat, S.; et~al.
\newblock 2023.
\newblock Gpt-4 technical report.
\newblock {\em arXiv preprint arXiv:2303.08774}.

\bibitem[\protect\citeauthoryear{Bommasani \bgroup et al\mbox.\egroup }{2021}]{bommasani2021opportunities}
Bommasani, R.; Hudson, D.~A.; Adeli, E.; Altman, R.; Arora, S.; von Arx, S.; Bernstein, M.~S.; Bohg, J.; Bosselut, A.; Brunskill, E.; et~al.
\newblock 2021.
\newblock On the opportunities and risks of foundation models.
\newblock {\em arXiv preprint arXiv:2108.07258}.

\bibitem[\protect\citeauthoryear{Chalkidis and Kampas}{2019}]{chalkidis2019deep}
Chalkidis, I., and Kampas, D.
\newblock 2019.
\newblock Deep learning in law: early adaptation and legal word embeddings trained on large corpora.
\newblock {\em Artificial Intelligence and Law} 27(2):171--198.

\bibitem[\protect\citeauthoryear{Gesnouin \bgroup et al\mbox.\egroup }{2024}]{gesnouin2024llamandement}
Gesnouin, J.; Tannier, Y.; Da~Silva, C.~G.; Tapory, H.; Brier, C.; Simon, H.; Rozenberg, R.; Woehrel, H.; Yakaabi, M.~E.; Binder, T.; et~al.
\newblock 2024.
\newblock Llamandement: Large language models for summarization of french legislative proposals.
\newblock {\em arXiv preprint arXiv:2401.16182}.

\bibitem[\protect\citeauthoryear{Goebel \bgroup et al\mbox.\egroup }{2023}]{goebel2023summary}
Goebel, R.; Kano, Y.; Kim, M.-Y.; Rabelo, J.; Satoh, K.; and Yoshioka, M.
\newblock 2023.
\newblock Summary of the competition on legal information, extraction/entailment (coliee) 2023.
\newblock In {\em Proceedings of the Nineteenth International Conference on Artificial Intelligence and Law},  472--480.

\bibitem[\protect\citeauthoryear{Huang \bgroup et al\mbox.\egroup }{2023}]{huang2023towards}
Huang, H.; Wu, S.; Liang, X.; Wang, B.; Shi, Y.; Wu, P.; Yang, M.; and Zhao, T.
\newblock 2023.
\newblock Towards making the most of llm for translation quality estimation.
\newblock In {\em CCF International Conference on Natural Language Processing and Chinese Computing},  375--386.
\newblock Springer.

\bibitem[\protect\citeauthoryear{Imani \bgroup et al\mbox.\egroup }{2019}]{imani2019deep}
Imani, A.; Vakili, A.; Montazer, A.; and Shakery, A.
\newblock 2019.
\newblock Deep neural networks for query expansion using word embeddings.
\newblock In {\em Advances in Information Retrieval: 41st European Conference on IR Research, ECIR 2019, Cologne, Germany, April 14--18, 2019, Proceedings, Part II 41},  203--210.
\newblock Springer.

\bibitem[\protect\citeauthoryear{Jagerman \bgroup et al\mbox.\egroup }{2023}]{jagerman2023query}
Jagerman, R.; Zhuang, H.; Qin, Z.; Wang, X.; and Bendersky, M.
\newblock 2023.
\newblock Query expansion by prompting large language models.
\newblock {\em arXiv preprint arXiv:2305.03653}.

\bibitem[\protect\citeauthoryear{Kayalvizhi, Thenmozhi, and Aravindan}{2019}]{kayalvizhi2019legal}
Kayalvizhi, S.; Thenmozhi, D.; and Aravindan, C.
\newblock 2019.
\newblock Legal assistance using word embeddings.
\newblock In {\em FIRE (Working Notes)},  36--39.

\bibitem[\protect\citeauthoryear{Kim \bgroup et al\mbox.\egroup }{2022}]{kim2022coliee}
Kim, M.-Y.; Rabelo, J.; Goebel, R.; Yoshioka, M.; Kano, Y.; and Satoh, K.
\newblock 2022.
\newblock Coliee 2022 summary: Methods for legal document retrieval and entailment.
\newblock In {\em JSAI International Symposium on Artificial Intelligence},  51--67.
\newblock Springer.

\bibitem[\protect\citeauthoryear{Kim, Rabelo, and Goebel}{2019}]{10.1145/3322640.3326742}
Kim, M.-Y.; Rabelo, J.; and Goebel, R.
\newblock 2019.
\newblock Statute law information retrieval and entailment.
\newblock In {\em Proceedings of the Seventeenth International Conference on Artificial Intelligence and Law}, ICAIL '19,  283–289.
\newblock New York, NY, USA: Association for Computing Machinery.

\bibitem[\protect\citeauthoryear{Laban \bgroup et al\mbox.\egroup }{2023}]{laban2023summedits}
Laban, P.; Kry{\'s}ci{\'n}ski, W.; Agarwal, D.; Fabbri, A.~R.; Xiong, C.; Joty, S.; and Wu, C.-S.
\newblock 2023.
\newblock Summedits: Measuring llm ability at factual reasoning through the lens of summarization.
\newblock In {\em Proceedings of the 2023 Conference on Empirical Methods in Natural Language Processing},  9662--9676.

\bibitem[\protect\citeauthoryear{Lai \bgroup et al\mbox.\egroup }{2023}]{lai2023large}
Lai, J.; Gan, W.; Wu, J.; Qi, Z.; and Yu, P.~S.
\newblock 2023.
\newblock Large language models in law: A survey.
\newblock {\em arXiv preprint arXiv:2312.03718}.

\bibitem[\protect\citeauthoryear{Landthaler \bgroup et al\mbox.\egroup }{2016}]{landthaler2016extending}
Landthaler, J.; Waltl, B.; Holl, P.; and Matthes, F.
\newblock 2016.
\newblock Extending full text search for legal document collections using word embeddings.
\newblock In {\em JURIX},  73--82.

\bibitem[\protect\citeauthoryear{Louis and Spanakis}{2022}]{louis2022statutory}
Louis, A., and Spanakis, G.
\newblock 2022.
\newblock A statutory article retrieval dataset in french.
\newblock In {\em Proceedings of the 60th Annual Meeting of the Association for Computational Linguistics (Volume 1: Long Papers)},  6789--6803.

\bibitem[\protect\citeauthoryear{Nguyen \bgroup et al\mbox.\egroup }{2016}]{DBLP:conf/nips/NguyenRSGTMD16}
Nguyen, T.; Rosenberg, M.; Song, X.; Gao, J.; Tiwary, S.; Majumder, R.; and Deng, L.
\newblock 2016.
\newblock {MS} {MARCO:} {A} human generated machine reading comprehension dataset.
\newblock In Besold, T.~R.; Bordes, A.; d'Avila Garcez, A.~S.; and Wayne, G., eds., {\em Proceedings of the Workshop on Cognitive Computation: Integrating neural and symbolic approaches 2016 co-located with the 30th Annual Conference on Neural Information Processing Systems {(NIPS} 2016), Barcelona, Spain, December 9, 2016}, volume 1773 of {\em {CEUR} Workshop Proceedings}.
\newblock CEUR-WS.org.

\bibitem[\protect\citeauthoryear{Nguyen \bgroup et al\mbox.\egroup }{2023}]{nguyen2023nowj1}
Nguyen, T.-M.; Nguyen, X.-H.; Mai, N.-D.; Hoang, M.-Q.; Nguyen, V.-H.; Nguyen, H.-V.; Nguyen, H.-T.; and Vuong, T.-H.-Y.
\newblock 2023.
\newblock Nowj1@ alqac 2023: Enhancing legal task performance with classic statistical models and pre-trained language models.
\newblock {\em arXiv preprint arXiv:2309.09070}.

\bibitem[\protect\citeauthoryear{Nguyen \bgroup et al\mbox.\egroup }{2024}]{nguyen2024captain}
Nguyen, C.; Nguyen, P.; Tran, T.; Nguyen, D.; Trieu, A.; Pham, T.; Dang, A.; and Nguyen, L.-M.
\newblock 2024.
\newblock Captain at coliee 2023: Efficient methods for legal information retrieval and entailment tasks.
\newblock {\em arXiv preprint arXiv:2401.03551}.

\bibitem[\protect\citeauthoryear{Pont \bgroup et al\mbox.\egroup }{2023}]{pont2023legal}
Pont, T.~D.; Galli, F.; Loreggia, A.; Pisano, G.; Rovatti, R.; and Sartor, G.
\newblock 2023.
\newblock Legal summarisation through llms: The prodigit project.
\newblock {\em arXiv preprint arXiv:2308.04416}.

\bibitem[\protect\citeauthoryear{Rane, Choudhary, and Rane}{2024}]{rane2024gemini}
Rane, N.; Choudhary, S.; and Rane, J.
\newblock 2024.
\newblock Gemini versus chatgpt: Applications, performance, architecture, capabilities, and implementation.
\newblock {\em Performance, Architecture, Capabilities, and Implementation (February 13, 2024)}.

\bibitem[\protect\citeauthoryear{Robertson, Zaragoza, and others}{2009}]{robertson2009probabilistic}
Robertson, S.; Zaragoza, H.; et~al.
\newblock 2009.
\newblock The probabilistic relevance framework: Bm25 and beyond.
\newblock {\em Foundations and Trends{\textregistered} in Information Retrieval} 3(4):333--389.

\bibitem[\protect\citeauthoryear{Sugathadasa \bgroup et al\mbox.\egroup }{2019}]{sugathadasa2019legal}
Sugathadasa, K.; Ayesha, B.; de~Silva, N.; Perera, A.~S.; Jayawardana, V.; Lakmal, D.; and Perera, M.
\newblock 2019.
\newblock Legal document retrieval using document vector embeddings and deep learning.
\newblock In {\em Intelligent Computing: Proceedings of the 2018 Computing Conference, Volume 2},  160--175.
\newblock Springer.

\bibitem[\protect\citeauthoryear{Sun}{2023}]{sun2023short}
Sun, Z.
\newblock 2023.
\newblock A short survey of viewing large language models in legal aspect.
\newblock {\em arXiv preprint arXiv:2303.09136}.

\bibitem[\protect\citeauthoryear{Team \bgroup et al\mbox.\egroup }{2023}]{team2023gemini}
Team, G.; Anil, R.; Borgeaud, S.; Wu, Y.; Alayrac, J.-B.; Yu, J.; Soricut, R.; Schalkwyk, J.; Dai, A.~M.; Hauth, A.; et~al.
\newblock 2023.
\newblock Gemini: a family of highly capable multimodal models.
\newblock {\em arXiv preprint arXiv:2312.11805}.

\bibitem[\protect\citeauthoryear{Thakur \bgroup et al\mbox.\egroup }{2021}]{thakur2021beir}
Thakur, N.; Reimers, N.; R{\"u}ckl{\'e}, A.; Srivastava, A.; and Gurevych, I.
\newblock 2021.
\newblock {BEIR}: A heterogeneous benchmark for zero-shot evaluation of information retrieval models.
\newblock In {\em Thirty-fifth Conference on Neural Information Processing Systems Datasets and Benchmarks Track (Round 2)}.

\bibitem[\protect\citeauthoryear{Touvron \bgroup et al\mbox.\egroup }{2023}]{touvron2023llama}
Touvron, H.; Lavril, T.; Izacard, G.; Martinet, X.; Lachaux, M.-A.; Lacroix, T.; Rozi{\`e}re, B.; Goyal, N.; Hambro, E.; Azhar, F.; et~al.
\newblock 2023.
\newblock Llama: Open and efficient foundation language models.
\newblock {\em arXiv preprint arXiv:2302.13971}.

\bibitem[\protect\citeauthoryear{Tran, Nguyen, and Satoh}{2019}]{10.1145/3322640.3326740}
Tran, V.; Nguyen, M.~L.; and Satoh, K.
\newblock 2019.
\newblock Building legal case retrieval systems with lexical matching and summarization using a pre-trained phrase scoring model.
\newblock In {\em Proceedings of the Seventeenth International Conference on Artificial Intelligence and Law}, ICAIL '19,  275–282.
\newblock New York, NY, USA: Association for Computing Machinery.

\bibitem[\protect\citeauthoryear{Trautmann, Petrova, and Schilder}{2022}]{trautmann2022legal}
Trautmann, D.; Petrova, A.; and Schilder, F.
\newblock 2022.
\newblock Legal prompt engineering for multilingual legal judgement prediction.
\newblock {\em arXiv preprint arXiv:2212.02199}.

\bibitem[\protect\citeauthoryear{Wang \bgroup et al\mbox.\egroup }{2024}]{wang2024can}
Wang, S.; Wei, Z.; Choi, Y.; and Ren, X.
\newblock 2024.
\newblock Can llms reason with rules? logic scaffolding for stress-testing and improving llms.
\newblock {\em arXiv preprint arXiv:2402.11442}.

\bibitem[\protect\citeauthoryear{Wang, Yang, and Wei}{2023}]{wang2023query2doc}
Wang, L.; Yang, N.; and Wei, F.
\newblock 2023.
\newblock Query2doc: Query expansion with large language models.
\newblock In {\em The 2023 Conference on Empirical Methods in Natural Language Processing}.

\bibitem[\protect\citeauthoryear{Yang \bgroup et al\mbox.\egroup }{2023}]{yang2023harnessing}
Yang, J.; Jin, H.; Tang, R.; Han, X.; Feng, Q.; Jiang, H.; Zhong, S.; Yin, B.; and Hu, X.
\newblock 2023.
\newblock Harnessing the power of llms in practice: A survey on chatgpt and beyond.
\newblock {\em ACM Transactions on Knowledge Discovery from Data}.

\bibitem[\protect\citeauthoryear{Yao \bgroup et al\mbox.\egroup }{2023}]{yao2023empowering}
Yao, B.; Jiang, M.; Yang, D.; and Hu, J.
\newblock 2023.
\newblock Empowering llm-based machine translation with cultural awareness.
\newblock {\em arXiv preprint arXiv:2305.14328}.

\bibitem[\protect\citeauthoryear{Yu, Quartey, and Schilder}{2022}]{yu2022legal}
Yu, F.; Quartey, L.; and Schilder, F.
\newblock 2022.
\newblock Legal prompting: Teaching a language model to think like a lawyer.
\newblock {\em arXiv preprint arXiv:2212.01326}.

\bibitem[\protect\citeauthoryear{Zhao \bgroup et al\mbox.\egroup }{2023}]{zhao2023survey}
Zhao, W.~X.; Zhou, K.; Li, J.; Tang, T.; Wang, X.; Hou, Y.; Min, Y.; Zhang, B.; Zhang, J.; Dong, Z.; et~al.
\newblock 2023.
\newblock A survey of large language models.
\newblock {\em arXiv preprint arXiv:2303.18223}.

\bibitem[\protect\citeauthoryear{Zheng \bgroup et al\mbox.\egroup }{2020}]{zheng2020bert}
Zheng, Z.; Hui, K.; He, B.; Han, X.; Sun, L.; and Yates, A.
\newblock 2020.
\newblock Bert-qe: contextualized query expansion for document re-ranking.
\newblock {\em arXiv preprint arXiv:2009.07258}.

\bibitem[\protect\citeauthoryear{Zheng \bgroup et al\mbox.\egroup }{2021}]{10.1016/j.ipm.2021.102672}
Zheng, Z.; Hui, K.; He, B.; Han, X.; Sun, L.; and Yates, A.
\newblock 2021.
\newblock Contextualized query expansion via unsupervised chunk selection for text retrieval.
\newblock {\em Inf. Process. Manage.} 58(5).

\bibitem[\protect\citeauthoryear{Zhong \bgroup et al\mbox.\egroup }{2020}]{zhong2020does}
Zhong, H.; Xiao, C.; Tu, C.; Zhang, T.; Liu, Z.; and Sun, M.
\newblock 2020.
\newblock How does nlp benefit legal system: A summary of legal artificial intelligence.
\newblock In {\em Proceedings of the 58th Annual Meeting of the Association for Computational Linguistics},  5218--5230.

\bibitem[\protect\citeauthoryear{Zhou, Huang, and Wu}{2023}]{zhou2023boosting}
Zhou, Y.; Huang, H.; and Wu, Z.
\newblock 2023.
\newblock Boosting legal case retrieval by query content selection with large language models.
\newblock In {\em Proceedings of the Annual International ACM SIGIR Conference on Research and Development in Information Retrieval in the Asia Pacific Region},  176--184.

\end{thebibliography}

\end{document}